\documentclass[10pt,twocolumn,letterpaper]{article}

\usepackage{cvpr}              

\usepackage{graphicx}
\usepackage{amsmath}
\usepackage{amssymb}
\usepackage{booktabs}
\usepackage{mathtools}
\usepackage[ruled,linesnumbered]{algorithm2e}
\usepackage{multirow}

\usepackage[pagebackref,breaklinks,colorlinks]{hyperref}

\usepackage[capitalize]{cleveref}
\crefname{section}{Sec.}{Secs.}
\Crefname{section}{Section}{Sections}
\Crefname{table}{Table}{Tables}
\crefname{table}{Tab.}{Tabs.}

\def\cvprPaperID{6361} 
\def\confName{CVPR}
\def\confYear{2022}

\begin{document}

\title{iSeg3D: An Interactive 3D Shape Segmentation Tool}

\author{Sucheng Qian, Liu Liu, Wenqiang Xu, Cewu Lu\\
Shanghai JiaoTong University \\
{\tt\small \{qiansucheng, liuliu1993, vinjohn, lucewu\}@sjtu.edu.cn}}

\maketitle

\begin{abstract}
    A large-scale dataset is essential for learning good features in 3D shape understanding, but there are only a few datasets that can satisfy deep learning training. One of the major reasons is that current tools for annotating per-point semantic labels using polygons or scribbles are tedious and inefficient. To facilitate segmentation annotations in 3D shapes, we propose an effective annotation tool, named \textbf{i}nteractive \textbf{Seg}mentation for \textbf{3D} shape (\textbf{iSeg3D}). It can obtain a satisfied segmentation result with minimal human clicks ($<$ 10). Under our observation, most objects can be considered as the composition of finite primitive shapes, and we train iSeg3D model on our built primitive-composed shape data to learn the geometric prior knowledge in a self-supervised manner. Given human interactions, the learned knowledge can be used to segment parts on arbitrary shapes, in which positive clicks help associate the primitives into the semantic parts and negative clicks can avoid over-segmentation. Besides, We also provide an online human-in-loop fine-tuning module that enables the model perform better segmentation with less clicks. Experiments demonstrate the effectiveness of iSeg3D on PartNet shape segmentation. Data and codes will be made publicly available.

\end{abstract}

\section{Introduction}
\label{sec:intro}

Learning to segment and understand each part of the 3D shape is essential for many computer vision or robotics applications, such as object affordance analysis \cite{hu2018predictive}, shape editing \cite{yu2004mesh}, shape generation \cite{mo2019structurenet}, human-object interactions \cite{kim2014shape2pose}, etc. Recent years have witnessed the success of deep learning based 3D shape segmentation frameworks \cite{luo2020learning,yi2017syncspeccnn,yi2019gspn,hou20193d}. These works are usually able to parse 3D objects into several parts in a supervised manner \cite{qi2017pointnet++,li2018pointcnn}, which are trained on large-scale datasets \cite{mo2019partnet}. This requires the rich part annotations from model repositories for synthesizing numerous dynamic training data. 

\begin{figure}[t!]
\centering 
\includegraphics[width=\linewidth]{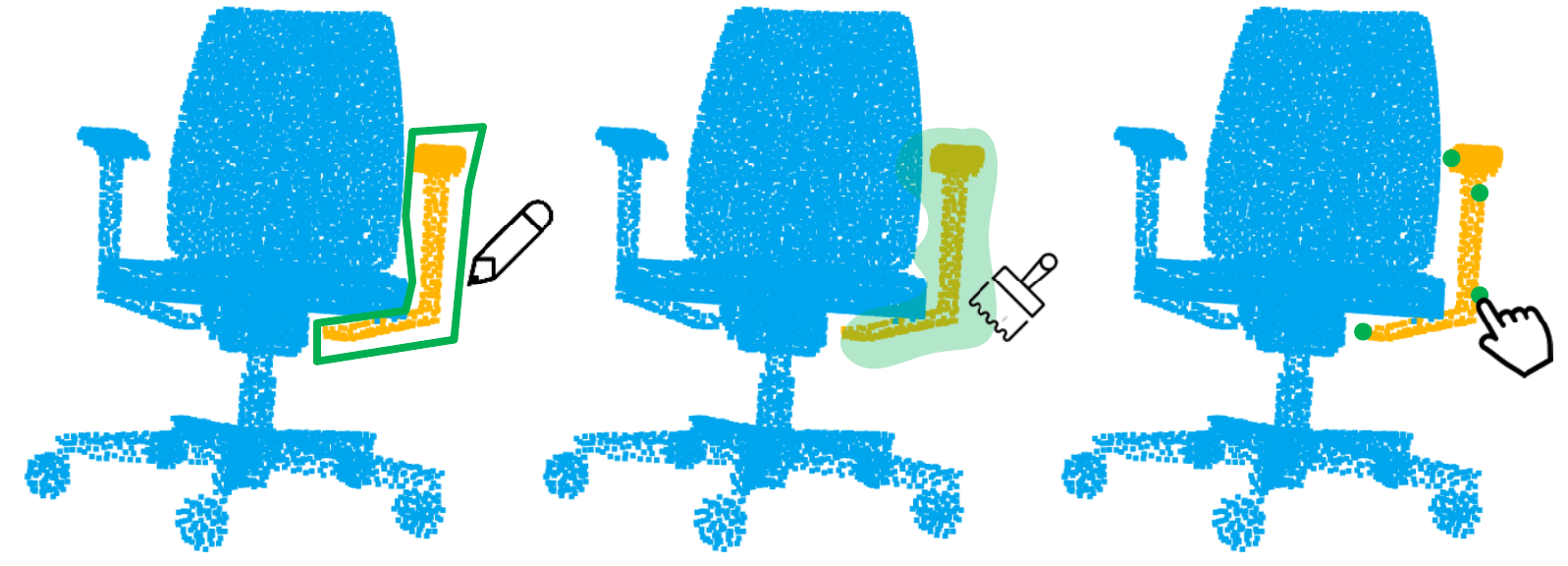}
\caption{Shape annotation methods. \textbf{Left:} Polygon based segmentation annotator. \textbf{Middle:} Scribbles based segmentation annotator. \textbf{Right:} Our iSeg3D annotation tool that requires only a few clicks ($<$ 10) from the user for shape segmentation.}
\label{fig:teaser}
\end{figure}

To build large-scale datasets with precise annotations, many works focus on designing shape labeling interfaces that employ UI workflows for operating on 3D models and collecting part instances \cite{yi2016scalable,mo2019partnet}. When segmenting each part region, these systems tend to provide polygons \cite{huang2018robust} or scribbles \cite{chang2018linking} for users to perform annotation on 2D renderings, as shown in Fig. \ref{fig:teaser}. However, these annotation workflows are limited to tedious human interactions required.


As well known, most of parts in 3D shape are similar to some primitive shapes (plane, cylinder, etc.) or can be composed by finite primitives \cite{nieuwenhuisen2012shape,yamanobe2010grasp,shu2016unsupervised,somani2014object}. 

Motivated by this, we attempt to enable the deep learning model to learn how to associate the geometric primitives into a part. To this end, we propose an effective annotation tool, named \textbf{i}nteractive \textbf{Seg}mentation for \textbf{3D} shape (\textbf{iSeg3D}), using interactive segmentation mechanism to provide segment annotation based on pror primitive knowledge. To build the training data, we construct a large-scale primitive repository that is extracted from the ABC shapes \cite{koch2019abc} and re-organize these primitives to build numerous training samples with automatically generated ground truths. Thus, the iSeg3D model can be trained on these primitive-composed shapes in a self-supervision manner, and the learned model can be easily used to infer arbitrary shapes for part segmentation interactively. In this way, users are only required to provide a few positive and negative label clicks ($<$10 clicks, 3-5 seconds for human operation) on each part region for good segmentation results.

In iSeg3D, given primitive-composed shapes, we first pre-train a point embedding space to learn the point-wise geometry-aware priors. Next, we use the primitive trained model to infer part segmentation on arbitrary shapes, where we propose a click-based iterative segmentation mechanism with human clicks. The clicks can be provided in two types: positive ones help associate primitives into the semantic parts and negative ones are used to refine the segments and avoid over-segmentation. We also propose two post-processing methods turn the model output into a connected, coherent segment. Besides, we provide an Online Human-in-Loop mechanism to learn the new primitive knowledge based on current clicks and it can help perform better segmentation results with less click numbers. Experiments on PartNet \cite{mo2019partnet} demonstrate that our iSeg3D can achieve efficient interactive segmentation performance on 3D shapes, which obtain \textbf{83.7\%} mean IoU, \textbf{7.7} and \textbf{9.0} for Number of Clicks (NoC) @mIoU 80 and 85.

The key contributions of our work can be summarized as follows:

\begin{figure*}[ht!]
\centering 
\includegraphics[width=\textwidth]{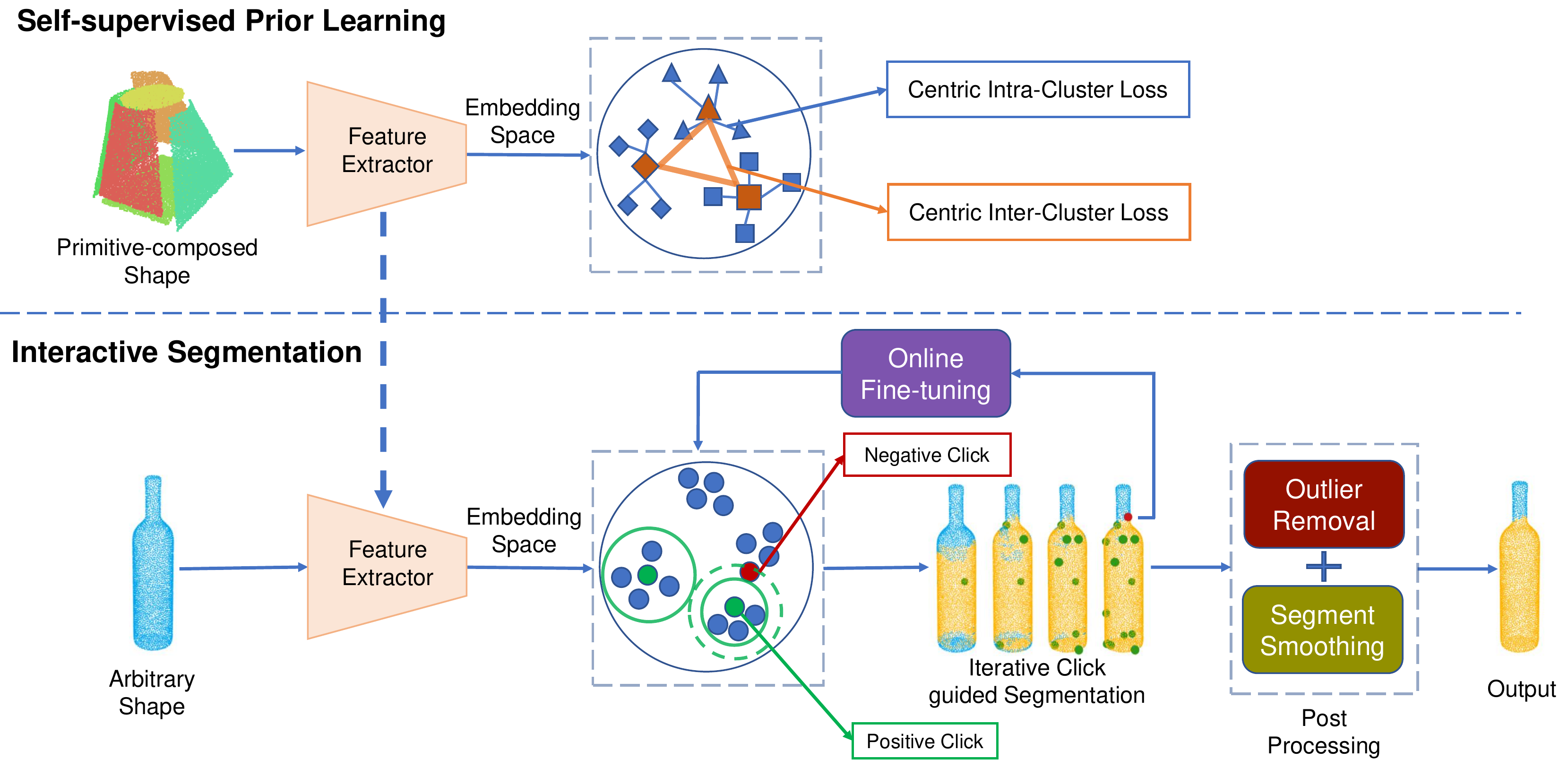}

\caption{Overview of our iSeg3D. We train our model in a self-supervision manner on the large-scale primitive-composed shapes. When used to segment an arbitrary shape, our model annotates the points sharing similar primitive-aware embedding with human positive clicks. Negative clicks are also provided for avoiding over-segmentation. Next, two post-processing algorithms are used to further refine the segments.
When a negative click occurs, we propose an online human-in-loop fine-tuning module to update the embedding space learning.}
\label{fig:architecture}
\end{figure*}

\begin{itemize}
    \item We propose an annotation tool iSeg3D that could be used as an effective 3D shape segmentation annotator, that exploits the primitives and learns the prior knowledge for interactive segmentation.
    
    
    
    \item Our iSeg3D can be easily used to infer part segments on arbitrary shapes by iterative human clicks. We propose two segmentation post-processing algorithms for segment refinement. Besides, we als provide an online learning module to learn new shape knowledge. 
    
    
    \item Experiments on PartNet demonstrate the effectiveness of our iSeg3D for shape segmentation, which obtain better performance compared with interactive segmentation baselines and supervised segmentation methods. Data and codes will be made publicly available. 
    
\end{itemize}

\section{Related Work}

\subsection{3D Shape Segmentation}

3D shape segmentation requires algorithms to precisely recognize the part-level regions, which is important for shape understanding. Most of the current works adopt supervised learning that learns a semantic feature vector for each point and group these points into finite parts \cite{mo2019partnet,kalogerakis2010learning,wang20183d}. Specifically, Guo et al. \cite{guo20153d} learns to label mesh faces with semantic labels defined by a human.

More recent works propose novel 3D deep network architectures segmenting shapes represented as 2D images \cite{kalogerakis20173d}, 3D voxels \cite{maturana20153d}, sparse volumetric representations \cite{klokov2017escape,riegler2017octnet} point clouds \cite{qi2017pointnet,wang2018sgpn,yi2019gspn} and graph-based representations \cite{yi2017syncspeccnn}. These methods take advantage of sufficient training samples of seen categories and demonstrate appealing performance for shape segmentation.

When transferred to unseen categories, several works use a bottom-up clustering mechanism that learns a per-pixel embedding and utilizes a clustering algorithm as a post-process to generate final segments \cite{wang2018sgpn,yi2019gspn,kaick2014shape,luo2020learning,hou20193d}. However, these methods might not perform well on unseen categories with only an average of 0.5 IoU. In this work, we aim to introduce interactive segmentation on 3D shapes and obtain a much better performance than current shape segmentation approaches. 

\subsection{Interactive Segmentation}

Interactive segmentation is a hot topic that has been studied for a long time. Classical methods tend to model the interactive segmentation task as an optimization problem. Given the user's clicks, some works represent the RGB image as a graph and find a globally optimal segmentation by introducing the energy function with the balanced region and boundary information \cite{boykov2001interactive}. Besides, to relieve the dependence of color space, GrabCut employs the Gaussian mixture model to find the connected component of the targeted object in the RGB image \cite{rother2004grabcut}. On the other hand, the geodesic path is also adopted to replace graph cut to avoid its boundary-length bias \cite{bai2007geodesic}, in which Price et al. \cite{price2010geodesic} explores the geodesics' relative strengths while Gulshan et al. \cite{gulshan2010geodesic} calculates the geodesic between clicked points and other pixels to predict segments with minimum energy cost. While these classical methods can improve the interactive experience, they struggle with complex foreground/background color distributions in images. With the era of deep learning, several works attempt to design convolutional neural networks to embed the user's clicks into distance maps \cite{xu2016deep}. To refine the predicted mask by human clicks, RIS-Net adds an extra local branch into FCN for refinement \cite{liew2017regional}. Furthermore, BRS \cite{jang2019interactive} and f-BRS \cite{sofiiuk2020f} use a backward propagation mechanism to fine-tune the guidance map in an online manner. To solve the generalization failure problem, Chen et al. \cite{chen2021conditional} proposes a conditional diffusion strategy for better segmentation.

Among these deep learning methods, Feng et al. \cite{feng2016interactive} tries to investigate interactive segmentation on RGB-D images but still requires strong color input. Apart from this, few works focus on the 3D shape interactive segmentation task. In this work, we present our iSeg3D to study this area.



\section{Method}

In this section, we describe our iSeg3D annotation tool for interactive 3D shape segmentation. In our method, we employ a self-supervised prior learning mechanism to enable the deep learning model to learn geometry-aware priors using our ABC Reshuffle data (Sec. \ref{sec:prior_learning}). During interactive segmentation, the input shape is processed by our iSeg3D for feature extraction and iteratively clustered into part segments based on human clicks (Sec. \ref{sec:interactive_seg}). After that, we introduce two post-processing methods to turn the predicted annotation mask into a connected, coherent part segment (Sec. \ref{sec:post_processing}). Furthermore, we also provide a fine-tuning mode that allows our iSeg3D to learn the new shape knowledge from human interactions online (Sec. \ref{sec:online_finetune}). The overall pipeline of iSeg3D is illustrated in Fig. \ref{fig:architecture}.

\subsection{Self-Supervised Prior Learning}\label{sec:prior_learning}

\begin{figure}
    \centering
    \includegraphics[width=\linewidth]{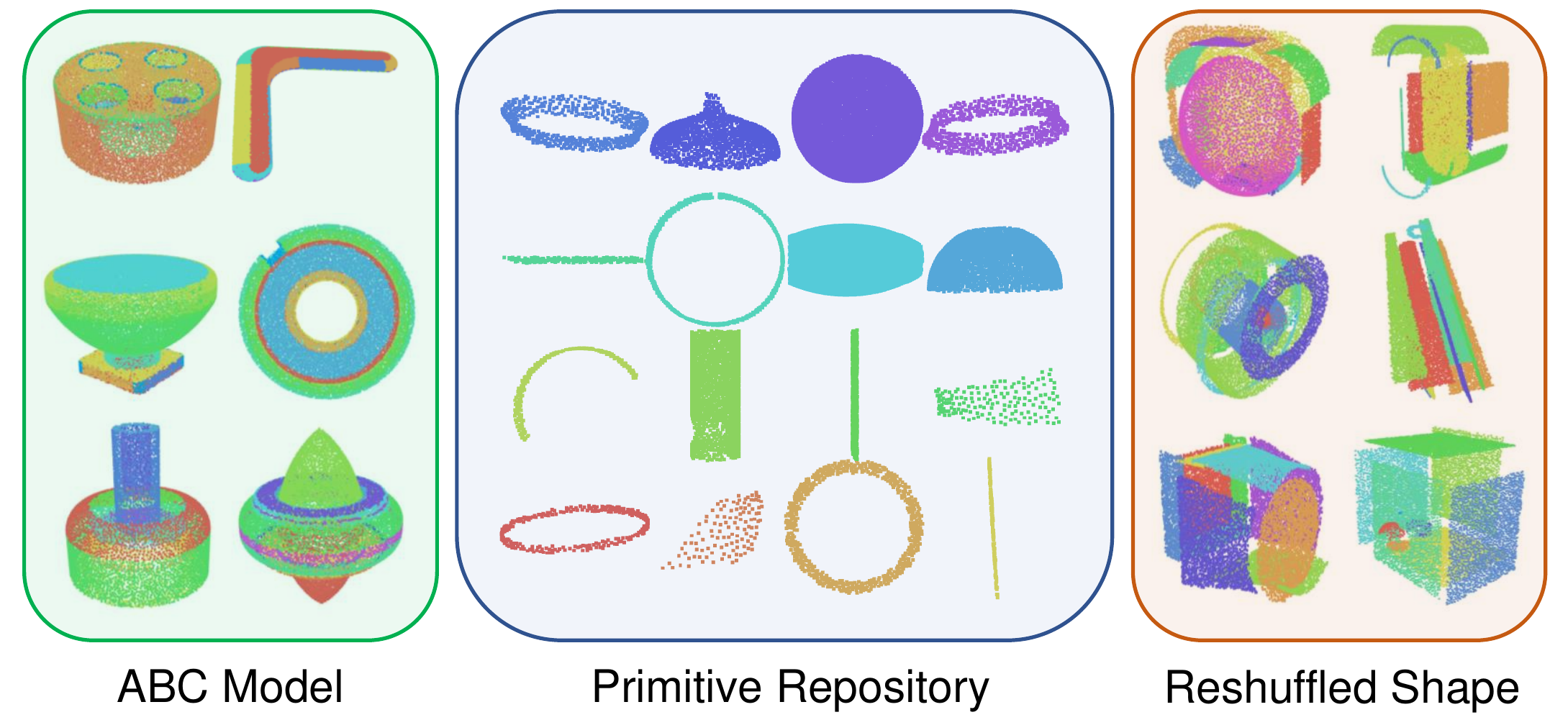}
    \caption{ABC Reshuffle dataset generation for primitive prior learning. We decompose the ABC models into primitives and re-organize them into new reshuffled shapes.}
    \label{fig:abc_reshuffle}
\end{figure}

\paragraph{ABC Reshuffle Data} We build our training data for iSeg3D to learn primitive knowledge in a self-supervision manner. Specifically, we extract she shape primitives (plane, circle, cylinder, etc.) from the ABC dataset \cite{koch2019abc} to construct a large-scale primitive repository. Then, we can automatically generate numerous training samples by randomly re-organizing these primitive shapes into new shapes, named ABC Reshuffle data. In this way, we can train our iSeg3D on the shapes with generated ground truths by self-supervision. The process of ABC Reshuffle is is illustrated in Fig. \ref{fig:abc_reshuffle}. For more details about how to assemble these primitives into shapes, please refer to supplementary materials. 


\paragraph{Prior Learning} We regard the prior primitive knowledge learning as enabling a deep learning model to learn point-level embeddings revealing the primitives composing the input shape, where points from the same patch are embedded closer and points from different patches form distinguishable clusters. 

The model input is a 3D point cloud with normal direction $P = \{ p_i \in \mathbb{R}^6 \vert i=1, \cdots, N \}$. The ground truth partitions point cloud $P$ into $K$ segments, in which each segment indicates one primitive in the primitive-composed shape. The set of point indices of the $i_{th}$ segment is denoted as index set $S_i$. Then we use PointNet++ \cite{qi2017pointnet++} backbone to extract point-wise feature embedding $Z= \{ z_i = F(p_i) \vert i=1,\cdots,N\}$ from $P$, where $z_i$ encodes the geometric contextual information extracted by model $F$ around point $p_i$. For learning the embedding space to distinguish these segments, we design the Centric Intra-Cluster loss $\mathcal{L}_{ctr}$ and Centric Inter-Cluster loss $\mathcal{L}_{cte}$ to train our model. 





\paragraph{Centric Intra-Cluster Loss} To correctly cluster primitive segments, we aim to learn a closer embedding space for the points from the same primitive. But for a point cloud, computing point-wise embedding distances is very time-consuming. In our method, we propose Centric Intra-Cluster loss $\mathcal{L}_{ctr}$ that aims to optimize the point-to-center offsets. In detail, we first extract the embedding center $\bar{z}_i$ for each primitive segment:

\begin{equation}
    \bar{z}_i = \frac{\sum_{j \in S_i} z_j}{\vert S_i \vert}
\end{equation}

\noindent Based on these embedding centers $\bar{Z}$, we define the Centeric Intra-Cluster loss $\mathcal{L}_{ctr}$:

\begin{equation}
    \mathcal{L}_{ctr} = \frac{1}{K} \sum_{i=1}^K [ 
        \frac{1}{\vert S_i \vert} \sum_{j \in S_i}  max \{\Vert z_j - \bar{z}_i \Vert_2 - \epsilon_{ctr}, 0\} 
    ]
\end{equation}

\noindent where $\epsilon_{ctr}$ is a margin threshold and the loss is calculated only when the embedding distance exceeds $\epsilon_{ctr}$. 




\paragraph{Centric Inter-Cluster Loss} We also define Centric Inter-Cluster loss $\mathcal{L}_{cte}$ that learns to separate the embedding centers of primitives away from each other. The loss function is defined as:


\begin{align}
    \mathcal{L}_{cte} & = \frac{1}{K (K-1)} \sum_{i=1}^k \sum_{j=1, i \neq j}^k max \{\epsilon_{cte} - \Vert \bar{z}_i - \bar{z}_j \Vert_2, 0 \} 
\end{align}

\noindent where we take the average over each pair of segments $\{S_i, S_j\}$. We also penalize the euclidean distance between two segments' embeddings $\{\bar{z}_i, \bar{z}_j\}$ if they are closer than a given threshold $\epsilon_{cte}$. 


\paragraph{The Overall Loss} Additionally, the average all point embedding norm as a regularization loss to keep the prior embedding space compact.

\begin{align}
    \mathcal{L}_{reg} & = \frac{1}{N} \sum_{i=1}^N ||z_i||_2
\end{align}

The overall loss is the weighted combination of $\mathcal{L}_{ctr}$, $\mathcal{L}_{cte}$, $\mathcal{L}_{reg}$. 

\begin{equation}
    \mathcal{L}  = \lambda_{ctr} \mathcal{L}_{ctr} + \lambda_{cte}\mathcal{L}_{cte} + \lambda_{reg}\mathcal{L}_{reg}
\end{equation}

\subsection{Iterative Segmentation Mechanism}\label{sec:interactive_seg}

\paragraph{Overview} To transfer our iSeg3D into unseen shapes for interactive segmentation, we design an iterative shape segmentation mechanism that exploits both positive clicks $C^{pos} = \{ c^{pos}_i \vert i=1, \cdots , N^{pos}\}$ and negatives clicks $C^{neg}=\{c^{neg}_i \vert i=1,\cdots,N^{neg}\}$ provided by user. The positive clicks indicate the region of interest for segmentation, while the negative clicks are used to refine the annotation. This interactive segmentation strategy adaptively combines precision with efficiency because the user can use negative clicks to precisely control the annotation boundary in the embedding space. This enables the system to produce increasingly accurate annotation given more annotation budget. 

\paragraph{Positive Click Segmentation} With user's positive clicks $C^{pos}$, we define an annotation function $A$ such that the annotated segment $M$ is given by $M=\{p_i \vert A(p_i)=1\}$:  

\begin{equation}
    A(p_i) =
    \begin{cases*}
      1 & if $\exists c^{pos}_t \in C^{pos}$, $||z_i - F(c^{pos}_t)||_2 < r^{pos}_t$ \\
      0 & otherwise
    \end{cases*}
\end{equation}

\noindent where $z_i = F(p_i)$, and $r^{pos}_t$ denotes the neighborhood radius of click $c^{pos}_t$. A point $p_i$ belongs to $M$ if $p_i$ lies in the neighborhood of some positive click $c^{pos}_t$, which is the euclidean epsilon ball with radius $r^{pos}_t$ in the embedding space. The points with their embeddings in the neighborhood of $c^{pos}_t$ are expected to share the same primitive shape with click $c^{pos}_t$ so that each positive click can annotate its nearby primitive shape. 






\paragraph{Negative Click Refinement} A larger radius $r^{pos}$ saves the annotator's workload but results in a coarse segmentation that easily propagates annotation to an unwanted area. A predefined fixed radius for all positive clicks may not well adapt to arbitrary 3D shapes. Therefore we enable the annotator to use negative clicks $C^{neg}$ to refine existing annotation by adjusting the radius $r^{pos}$ of related positive clicks. We define the radius $r^{pos}_t$ of positive click $c^{pos}_t$ as:

\begin{equation}
    r^{pos}_t = \min(\alpha, D_t),
\end{equation}

\noindent where $\alpha$ is a predefined default radius for positive clicks. $D_t$ is the minimal distance between the embedding of positive click $c^{pos}_t$ and all negative clicks $C^{neg}$, which is defined as:  

\begin{equation}
    D_t = \min_{i} || F(c^{pos}_t) - F(c^{neg}_i)||_2
\end{equation}

\noindent After updating the radius $r^{pos}_t$, we can refine the segment $M$ by applying annotation function $A$. 





\begin{figure}
    \centering
    \includegraphics[width=.9\linewidth]{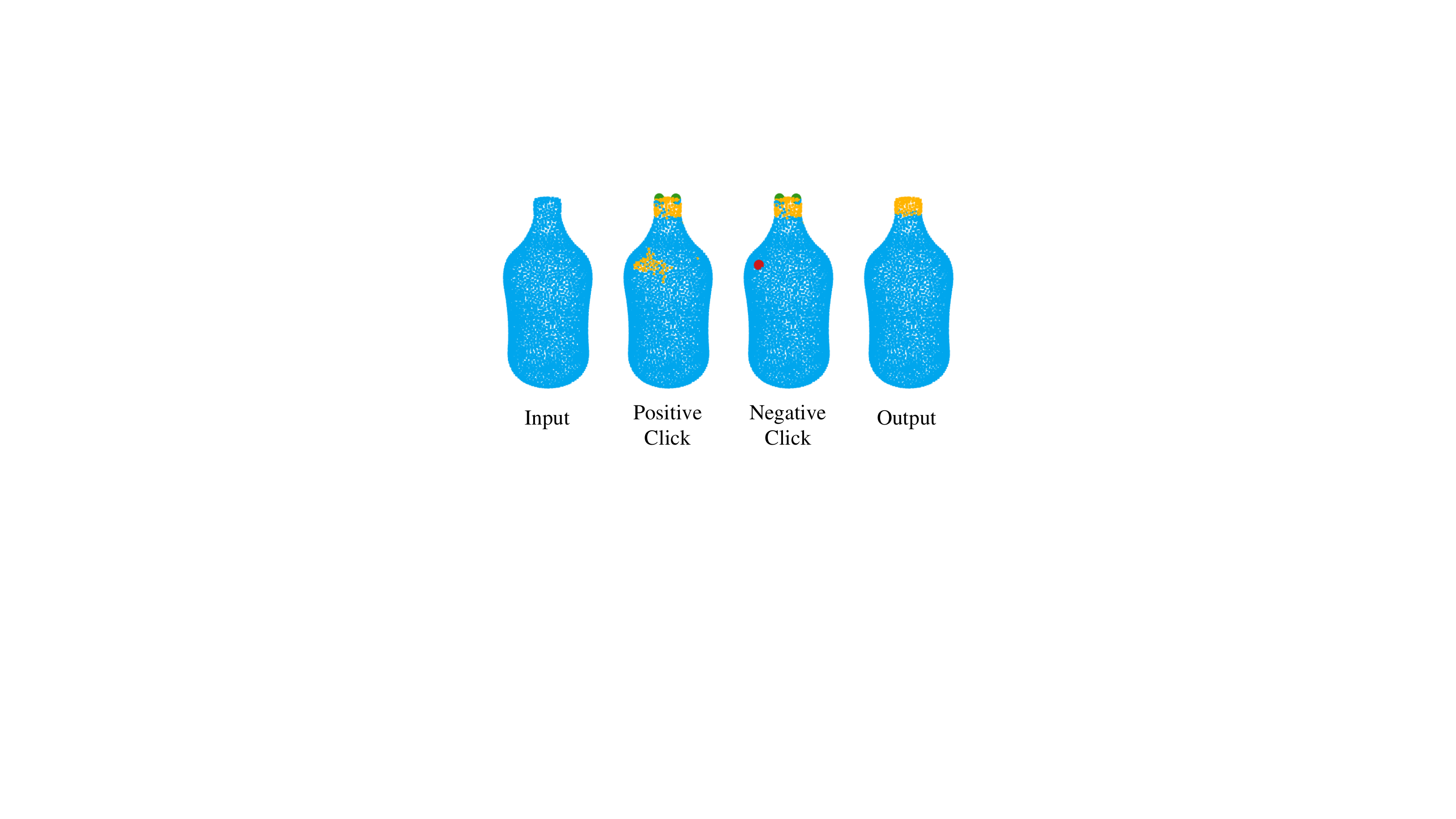}
    \caption{Iterative part segmentation generation with positive \& negative clicks.}
    \label{fig:interactive_ann}
\end{figure}

\subsection{Post-processing}\label{sec:post_processing}

In order to enforce a consistent topology between the point cloud euclidean space and the primitive prior embedding space, we propose two post-processing methods Outlier Removal and Segment Smoothing to regularize the annotated mask. 


\subsubsection{Outlier Removal}

Outlier Removal removes the annotated points that are close to the positive clicks in the embedding space but are not connected with them in the euclidean space. We can assume each instance part lies in a connected component. Thus for each annotated point, there should be some path on the segment connecting it to some positive click. 

We construct a graph $G=(V, E)$ to represent the point cloud $P$ with $V = P$. An undirected edge exists between point pair $(p_i, p_j)$ only if $p_j$ is the k-nearest neighbor of $p_i$ and $A(p_i) = A(p_j) = 1$. As is illustrated in Algorithm \ref{alg:outlier_removal}, we use Dijkstra algorithm to query for paths from each annotated point to positive clicks. If there's no path connecting the point to any positive click, then it is removed out of the annotation. 

\begin{algorithm}[tbh]
\caption{Outlier Removal}\label{alg:outlier_removal}
\SetKwInOut{Input}{input}
\SetKwInOut{Output}{output}

\Input{Point cloud $P$, annotated points $M$, positive clicks $C^{pos}$, graph vertex degree $N_{neighbor}$.}
\Output{Outlier removed annotated points $M^{(or)}$.}

Initialize graph $G=(V, E)$ with $V=P$ and $E=\emptyset$. 

\For{$p_i \in P$}{
    \For{$j \leftarrow 1$ \KwTo $N_{neighbor}$} {
        Let $p^{i}_j$ be $p_i$'s $j_{th}$ nearest neighbor in $P$. 
        
        \If{ $p_i \in M$ and $p^j_i \in M$}
        {Add undirected edge $(p_i, p^j_i)$ to $E$. 
        }
    }
}

Initialize $M^{(or)}$ by an empty set. 

\For{$M_i \in M$}{
    
    \If{
        \text{Exists some path from $M_i$ to any $c_j^{pos} \in C^{pos}$}}{
        Add $M_i$ into $M^{(or)}$. 
    }
}
return $M^{(or)}$

\end{algorithm}

\subsubsection{Segment Smoothing}

Segment Smoothing adds the points that are distant away from positive clicks in embedding space but are surrounded by the annotated ones in euclidean space. We use a simple smoothing method to make the annotated segment more geometrically coherent. As illustrated in Algorithm \ref{alg:segment_smooth}, for each unsegmented point, we query for its $N_{smooth}$ nearest neighboring points. If more than $\gamma$ percent of them are annotated, then we regard this point as also being annotated. This post-processing method is applied $N_{iter}$ times recurrently.

\begin{algorithm}[tbh]
\caption{Segment Smoothing}\label{alg:segment_smooth}
\SetKwInOut{Input}{input}\SetKwInOut{Output}{output}

\Input{Point cloud $P$, annotated point set $M$, neighborhood point number $N_{smooth}$.}

\Output{Smoothed segmentation mask $M^{(ss)}$.}

Initialize $M^{(ss)}$ as $M$. 

\For{$p_i \notin M$}{
    $B \leftarrow$ the set of $N_{smooth}$ nearest neighbors of point $p_j$ around the neighborhood.
    
    \If{$|L \cap B| / |B| > \gamma$}
    {
        Add $p_j$ into $M^{(ss)}$. 
    }
}

return $M^{(ss)}$
\end{algorithm}

\subsection{Online Human-in-Loop Fine-tuning}\label{sec:online_finetune}

During segmentation, we propose an online learning mechanism in a human-in-loop manner that allows the network to quickly learn primitive knowledge from unseen categories. Specifically, we fully exploit the human interactions of positive clicks and negative clicks to help our model fine-tune the parameters. Given user annotated positive clicks $C^{pos}$ and negative clicks $C^{neg}$, online fine-tuning energy is defined to minimize the affinity between each negative click embedding and its closest positive click embedding if the affinity is smaller than a threshold $\epsilon_{cte}$. The optimization function $E$ for fine-tuning the model is defined as:


\begin{equation}
    E = \frac{1}{N^{neg}} \sum_{i=1}^{N^{neg}} \max(\epsilon_{cte} - D_i, 0)
\end{equation}

\noindent where $D_i$ is the smallest distance between the embedding of negative click $c^{neg}_i$ and all positive clicks $C^{pos}$, which is defined as: 

\begin{equation}
    D_i = \min_j || F(c_i^{neg}) - F(c_j^{pos}) ||_2
\end{equation}


Besides, we also allow the user to use a scribble tool that can densely mark a set of negative clicks in the mislabeled area. This provides richer information for the optimization and costs little additional human effort compared with clicks.

\section{Experiments}



\subsection{Experimental Setup}

\paragraph{Implementation Details and Evaluation} We use PointNet++ \cite{qi2017pointnet++} as feature extractor in our iSeg3D. The input point clouds are sampled into 4,096 points. Besides, we use Adam as optimizer. Initial learning rate is 0.001 and the learning rate decay strategy is Cosine Annealing. The total training epoch is 1,000. The detailed hyper-parameters are: $\lambda_{ctr}=1$, $\lambda_{cte}=1$, $\lambda_{reg}=0.0001$, $\epsilon_{ctr}=0.25$, $\epsilon_{cte}=0.75$, $\alpha=0.35$, $\gamma=0.7, N_{neighbor}=3, N_{iter}=5, N_{smooth}=32$. For evaluation, We report per-part segmentation results using Intersection-over-Union (IoU), Number of Clicks (NoC) at IoU 80 and 85 metrics on 8 categories of PartNet test set. We use the segmentation performance of 10 user clicks as our results.

\paragraph{Click Simulation} We use an interactive simulation algorithm to approximate the annotator's behavior. Similar to an annotator with an intention to segment a target part of a 3D shape, the interactive simulation is provided with the ground truth target segment and aims to maximize IoU between the annotated segment and the ground truth. Given existing clicks $C^{pos}$ and $C^{neg}$, the algorithm greedily selects the click leading to the greatest IoU improvement in order to approximate the annotation behavior adopted by a rational annotator.

\paragraph{Baselines} Since there is no existing interactive segmentation methods on 3D shapes, we compare our approach with our defined interactive segmentation baselines. Furthermore, we also report the comparison with supervised segmentation baselines that are trained on the dataset containing seen categories.

\begin{itemize}
    \item \textbf{Interactive Segmentation Baselines. }We introduce two types of interactive segmentation algorithms: (1) Using classic \textbf{Fast Point Feature Histograms (FPFH)} \cite{rusu2009fast} for feature extraction, and then we exploit the FPFH embeddings into our iterative segmentation mechanism to generate part segmentation. (2) \textbf{Heatmap-based interactive segmentation} that is revised from a deep learning method \cite{mahadevan2018iteratively} where the user clicks are encoded as a heatmap smoothed by a Gaussian kernel.  This heatmap is then concatenated with the input shape and becomes an additional feature channel. A convolutional neural network takes this user annotated image and outputs the segmentation result by a standard pixel-wise classification.
    
    \item \textbf{Supervised Segmentation Baselines. }We also compare our approach with the supervised part instance segmentation methods. We follow the standard train set provided by PartNet, and train an instance segmentation model followed by Mo et al. \cite{mo2019partnet} as well as SGPN \cite{wang2018sgpn} model to predict the per-part segment.
\end{itemize}

\subsection{Performance on PartNet}


\begin{table*}[tbh]
    \centering
    \resizebox{\linewidth}{!}{
    \begin{tabular}{l|ccccccccc}
    \hline 
    
    
    Category & Bottle & Chair & Display & Faucet & Laptop & Mug & Refrigerator & Table & mean \\ 
    
    \hline 
    
        
    metric & \multicolumn{8}{c}{Per-category mean Intersection-over-Union (mIoU) using 10 clicks $\uparrow$} \\ 
    
    \hline
    \textit{Supervised Method} \\
    SGPN \cite{wang2018sgpn} & 72.7 & 80.5 & 85.4 & 66.8 & 91.2 & 82.8 & 52.5 & 79.4 & 76.4 \\ 
    Mo et al. \cite{mo2019partnet} & 77.9 & \textbf{85.1} & \textbf{92.4} & 70.9 & 94.5 & 86.2 & 57.4 & \textbf{84.4} & 81.1 \\
    \hline
    \textit{Interactive Method} \\
    FPFH-based \cite{rusu2009fast} & 53.8 & 55.6 & 48.5 & 52.8 & 70.4 & 60.2 & 37.5 & 38.3 & 52.1 \\ 
    Heatmap-based \cite{mahadevan2018iteratively} & 69.2 & 70.0 & 58.7 & 61.2 & 72.7 & 66.3 & 49.1 & 44.3 & 61.4 \\
    iSeg3D & \textbf{82.2} & 83.6 & 83.5 & \textbf{82.0} & \textbf{95.7} & \textbf{86.5} & \textbf{76.0} & 80.1 & \textbf{83.7} \\  
    
    \hline 

    metric & \multicolumn{8}{c}{Number of Clicks @80 per-category mIoU $\downarrow$} \\ 
    
    \hline 
    \textit{Interactive Method} \\
    FPFH-based \cite{rusu2009fast} & - & - & - & - & 10.2 & - & - & - & - \\ 
    Heatmap-based \cite{mahadevan2018iteratively} & 10.6 & 9.4 & 12.4 & 11.0 & 9.8 & 11.7 & 14.6 & 14.1 & 11.7\\  
    iSeg3D & \textbf{7.6} & \textbf{8.1} & \textbf{8.3} & \textbf{8.5} & \textbf{3.4} & \textbf{8.2} & \textbf{9.8} & \textbf{8.0} & \textbf{7.7} \\ 
    
    \hline 
    
    metric & \multicolumn{8}{c}{Number of Clicks @85 per-category mIoU $\downarrow$} \\ 
    
    \hline 
    \textit{Interactive Method} \\
    FPFH-based \cite{rusu2009fast} & - & - & - & - & 12.5 & - & - & - & - \\ 
    Heatmap-based \cite{mahadevan2018iteratively} & 11.9 & 10.5 & 13.5 & 12.2 & 11.6 & 12.7 & 14.9 & 14.5 & 12.7 \\ 
    iSeg3D & \textbf{8.9} & \textbf{9.6} & \textbf{9.8} & \textbf{9.8} & \textbf{4.2} & \textbf{9.3} & \textbf{11.0} & \textbf{9.0} & \textbf{9.0} \\ 
    \hline 
    
    \end{tabular}
    }
    \caption{Segmentation results on PartNet. We set the maximum click number 15. Note that we do not report NoC@80 and NoC85 for FPFH-based method in some categories since it cannot achieve 80\% and 85\% mIoU within 15 clicks. $\downarrow$ means the lower the better. $\uparrow$ means the higher the better.}
    \label{tab:main-results}
\end{table*}



In Table \ref{tab:main-results}, we report the performances of iSeg3D, supervised and interactive baselines on PartNet test set. 

First, we compare the final segmentation results by per-category mean IoU (mIoU) given 10 user clicks. Our iSeg3D achieves better performance compared with interactive segmentation baselines by a large margin across all categories, outperforming the FPFH-based method \cite{rusu2009fast} by \textbf{31.6\%} and the Heatmap-based method \cite{mahadevan2018iteratively} by \textbf{22.3\%} on average. Our approach also outperforms supervised baselines SGPN \cite{wang2018sgpn} by \textbf{7.3\%} and Mo et al. \cite{mo2019partnet} by \textbf{2.6\%}. Note that our interactive approach does not produce degenerated results as supervised methods do in categories with limited training data, such as Faucet and Refrigerator. Besides, our approach can achieve comparable results in the categories where supervised methods do not overfit. 

We also evaluate the interactive segmentation methods in the number of clicks required to achieve 80\% and 85\% mIoU. Our method can use \textbf{7.7}, \textbf{9.0} clicks on average to achieve 80\% and 85\% mIoU. While FPFN-based baseline fails to reach 80\% mIoU with a limit of 15 clicks, except for the Laptop because of its simple shape structure. The heatmap-based baseline uses 4.0 and 3.7 clicks more than ours on average to reach 80\% and 85\% mIoU, because it only propagates user clicks to the local geometric neighborhood instead of recognizing the underlying primitive shapes. 





\begin{figure*}[thb]
    \centering
    \includegraphics[width=\textwidth]{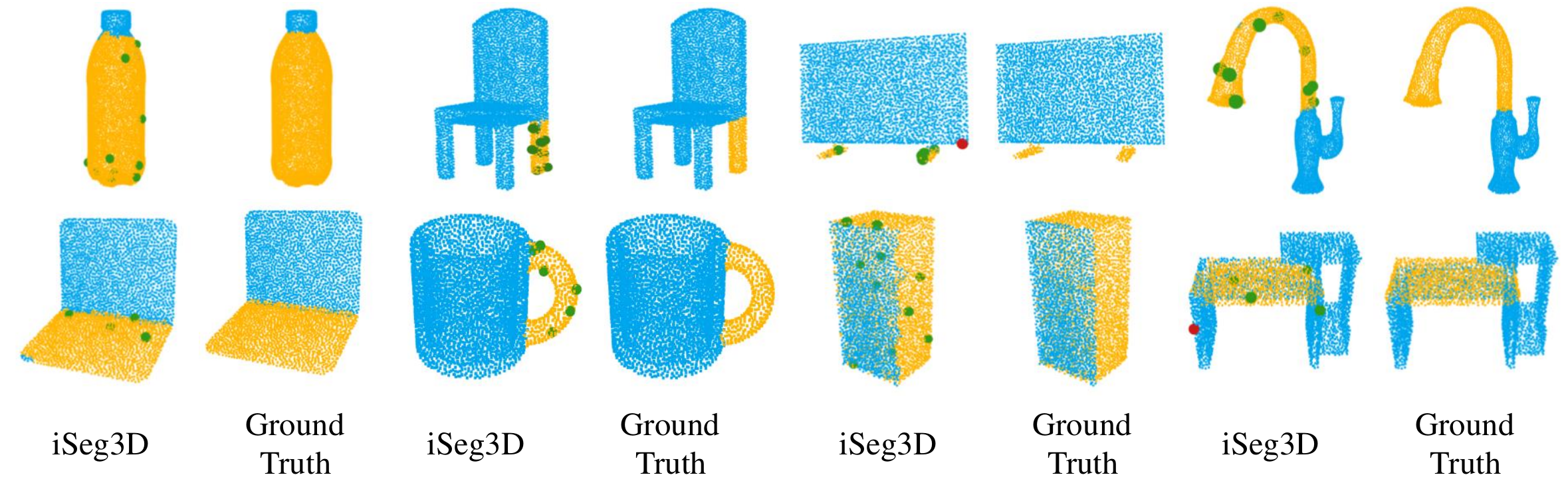}
    \caption{Qualitative Results. Here we demonstrate the segmentation result on only one part for each shape for best view.}
    \label{fig:qualitative_results}
\end{figure*}

\subsection{Ablation Studies}

\paragraph{Effect of Training Data Size} In Table \ref{tab:ablation_datasize}, we investigate the effectiveness of training data size. We randomly generate 10K, 20K, 30K and 40K primitive-composed shapes with re-organization from the primitive repository. As it can be seen, iSeg3D largely relies on the number of training samples, which can obtain \textbf{83.7\%} mIoU when training on 40K reshuffled shapes. Besides, it also only requires averagely \textbf{7.7} and \textbf{9.0} clicks to reach 80\% and 85\% mIoU. This could be explained by that larger primitive-composed shapes can provide enough prior knowledge for self-supervised learning, and more comprehensive geometric primitive information trained on a larger dataset leads to better segmentation results. On the other hand, when adding the training size from 30K to 40K, the segmentation mIoU only increases by \textbf{1.8\%}, which is lower than \textbf{3.9\%} that for adding 10K training samples to 20K. This is because the performance of our iSeg3D starts to converge on training size. Thus, we can conclude that 40K training samples are large enough for our model to learn primitive geometric knowledge.



\begin{table}[tbh]
    \centering
    \resizebox{\linewidth}{!}{
    \begin{tabular}{l|c|c|ccc}
    
    \hline 
    
        
    & \multicolumn{2}{c|}{Training Size} & mIoU $\uparrow$ & NoC@80 $\downarrow$ & NoC@85 $\downarrow$ \\ 
    
    \hline 
    
    \multirow{4}{*}{iSeg3D} & \multicolumn{2}{c|}{10K} & 73.4 & 11.7 & 13.8 \\ 
     & \multicolumn{2}{c|}{20K} & 77.3 & 9.5 & 11.4 \\ 
     & \multicolumn{2}{c|}{30K} & 81.9 & 8.1 & 9.6 \\ 
     & \multicolumn{2}{c|}{40K} & \textbf{83.7} & \textbf{7.7} & \textbf{9.0} \\ 
 
    \hline 
    
    \end{tabular}
    }
    \caption{Ablation Studies for the Training Dataset Size.}
    \label{tab:ablation_datasize}
\end{table}

\paragraph{Effect of Click Number}

\begin{figure}[t!]
    \centering 
    \includegraphics[width=\linewidth]{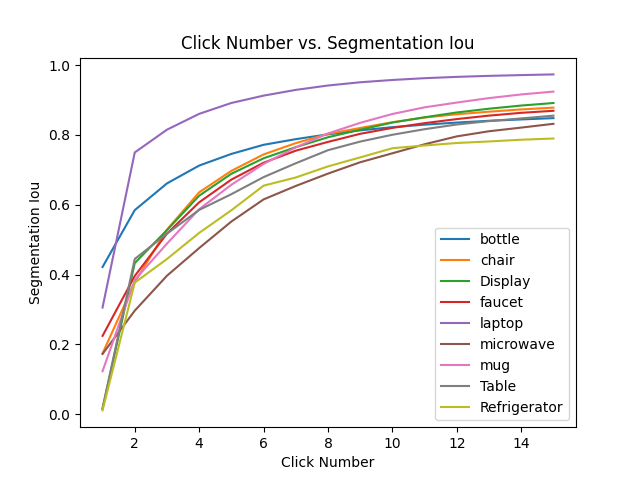}
    \caption{Effect of Click Number in Interactive Segmentation. }
    \label{fig:click-IoU}
\end{figure}

We study how the segmentation performance changes during the annotation process with different numbers of user clicks. Fig. \ref{fig:click-IoU} shows per-category mIoU of each category w.r.t. different click numbers ranging from 1 to 15. The segmentation results improve with more user clicks in general. Our approach achieves \textbf{62.3\%} mIoU with just 4 clicks, showing that the model learns the underlying primitive geometry so that the segment can be inferred by the sparse clicks upon it. Our approach outputs \textbf{83.7\%} mIoU given 10 clicks, producing acceptable segmentation results given minimal annotation workload in practice. The annotation performance further increases given more clicks, showing that our approach supports detailed segmentation refinement. 

\paragraph{Effect of Refinement Components}

We report the ablation study results of online refinement and post-processing in Table \ref{tab:ablation-postprocessing}. We observe that annotation performance is consistently improved with Outlier Removal and Segment Smoothing. The results illustrate that constraining the invariant topology between primitive prior embedding space and point cloud euclidean space produces better annotation results. Besides, the online refinement module also contributes to improvement in annotation performance, showing that online optimization helps adapt the prior embedding space to target unseen categories. 


\begin{table}[tbh]
    \centering
    \resizebox{\linewidth}{!}{
    \begin{tabular}{l|c|c|c|ccc}
    
    \hline 
    
        
    & OF & OR & SS & mIoU $\uparrow$ & NoC@80 $\downarrow$ & NoC@85 $\downarrow$ \\ 

    \hline 
    
    \multirow{4}{*}{iSeg3D} & \checkmark & & & 77.5 & 10.0 & 11.4 \\ 
     & \checkmark & \checkmark & & 78.7 & 9.2 & 10.8 \\ 
     & & \checkmark & \checkmark & 82.4 & 8.1 & 9.7 \\ 
     & \checkmark & \checkmark & \checkmark & \textbf{83.8} & \textbf{7.6} & \textbf{8.9} \\ 
 
    \hline 
    
    \end{tabular}
    }
    \caption{Ablation Studies for refinement components. \textbf{OF: }Online Fine-tuning Module. \textbf{OR: }Outlier Removal algorithm. \textbf{SS: }Segment Smoothing algorithm.}
    \label{tab:ablation-postprocessing}
\end{table}

\subsection{Failure Cases and Future Work}

\begin{figure}
    \centering
    \includegraphics[width=0.8\linewidth]{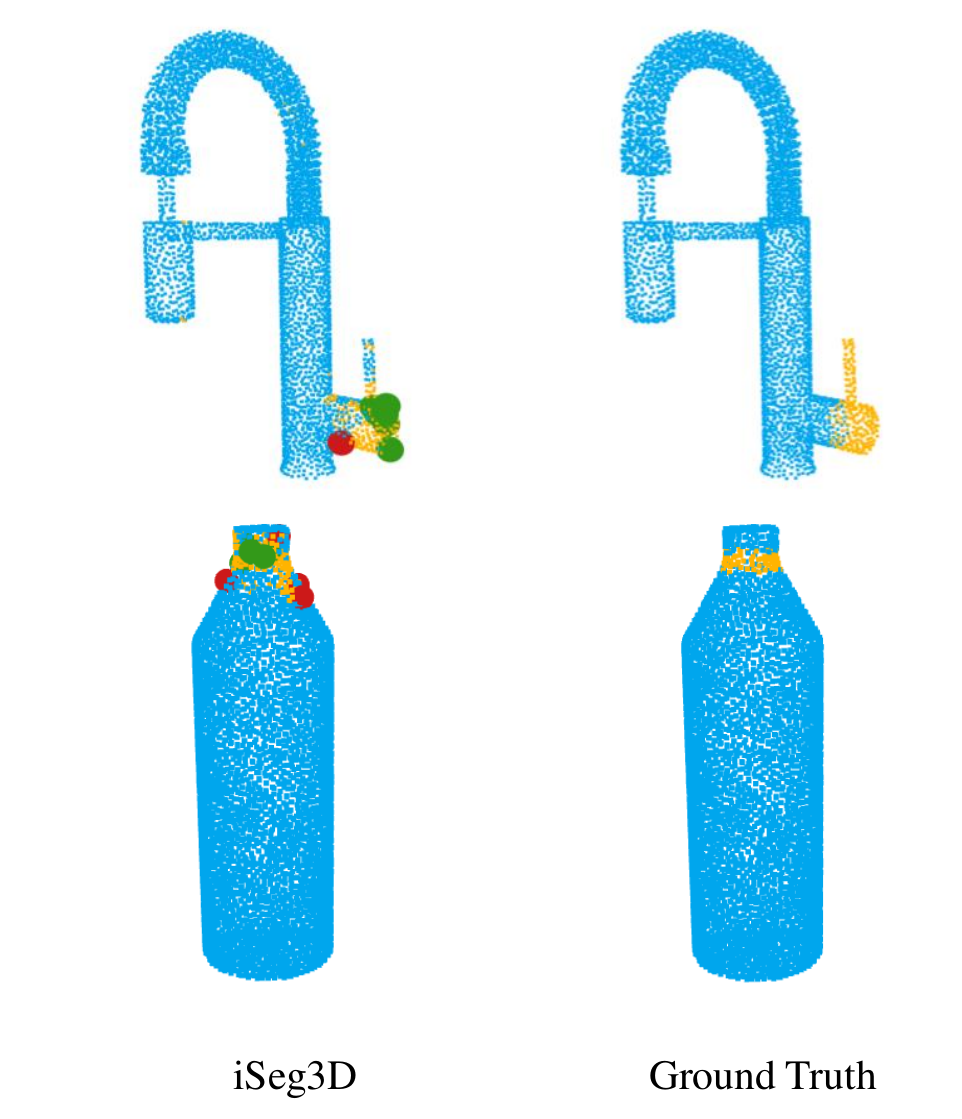}
    \caption{Failure cases of our iSeg3D method.}
    \label{fig:failure_cases}
\end{figure}

We visualize some of the failure cases in Fig. \ref{fig:failure_cases}. We observe that most of the segmentation failures lie on small-size part segmentation. This can be summarized into the following reasons: (1) some small parts (\eg faucet handle) are composed by slightly deformed primitives, our model is more difficult to learn the standard primitive that can fit them well compared with the large parts (\eg bottle). (2) When segmenting the part with a small size, the positive clicks provided by the user might be too concentrated, leading to redundant and inefficient human interactions. Future work will pay attention to providing primitive proposals for the user and automatically segment parts by user selection. 



\section{Conclusion}

In this paper, we present iSeg3D, an interactive 3D shape segmentation tool for click-based part annotation. Our iSeg3D can fully exploit the geometric knowledge and learn the priors from primitive shapes in a self-supervision manner. During segmentation, we allow the user to provide positive as well as negative clicks to iteratively refine the segments. We also propose an online fine-tuning module where the model can learn the knowledge from new categories by human-in-loop manner. Experiments demonstrate the effectiveness of our method in annotating shape segmentation. We hope iSeg3D can be an effective tool for building large-scale datasets and contributing to 3D vision community. 

{\small
\bibliographystyle{ieee_fullname}
\bibliography{egbib}
}

\end{document}